\documentclass[11pt]{article}
\usepackage{acl}
\usepackage{times}
\usepackage{latexsym}
\usepackage[T1]{fontenc}
\usepackage[utf8]{inputenc}
\usepackage{microtype}
\usepackage{inconsolata}
\usepackage{graphicx}
\usepackage{amsmath}
\usepackage{amssymb}
\usepackage{booktabs}
\usepackage{algorithm}
\usepackage{algpseudocode}
\usepackage{xcolor}
\usepackage{multirow}
\usepackage{array}
\usepackage{makecell}
\usepackage{tcolorbox}
\usepackage{enumitem}
\usepackage{colortbl}
\usepackage{listings}
\usepackage{pgfplots}
\usepackage{pgfplotstable}
\usepackage{tabularx}
\usepackage{float}
\usepackage{enumitem}
\tcbuselibrary{skins, breakable}

\title{Don't Lose Entities from Retrieval to Generation:\\Dual Entity Recovery RAG for multi-hop QA}

\author{
  Heechang Lee \\
  UNIST \\
  \texttt{heechang@unist.ac.kr} \\\And
  Dong-Young Lim\textsuperscript{\textdagger} \\
  UNIST \\
  \texttt{dlim@unist.ac.kr} \\
}

\begin{document}
\maketitle

\begingroup
\renewcommand{\thefootnote}{\fnsymbol{footnote}}
\footnotetext[2]{Corresponding author.}
\endgroup

\begin{abstract}
Retrieval-augmented multi-hop question answering (QA) decomposes a query into sub-questions and decomposes the corpus into smaller retrieval units such as sentences. Both forms of decomposition improve the pipeline, but we show that both share the same vulnerability, the loss of entity information, and that this loss breaks the pipeline at two separate points. The first point is retrieval, where a sub-question loses the entity resolved at the previous hop, leaving the retriever with nothing to match against. The second point is harder to see, because retrieval still appears to succeed. Once a passage is split into sentences, an isolated sentence loses the context that grounds its pronouns, so even with the correct sentence in hand the LLM cannot tell which entity the sentence is about. We isolate this second point as a distinct failure mode that we call lost-in-generation, and a retrieval-controlled experiment shows that it degrades answers even when the gold evidence is fixed in the context. We then propose Dual Entity Recovery RAG (DER-RAG), which keeps the grounding entity explicit from retrieval through to generation with two lightweight components, a two-way query decomposition that carries the resolved entity across sub-questions and a subject entity prefix attached to each sentence at generation time. DER-RAG needs no graph construction, no corpus modification, and no fine-tuning, yet on three multi-hop QA benchmarks it matches or exceeds strong baselines, including graph-based methods that depend on costly offline structures.
\end{abstract}

\section{Introduction}
\label{sec:intro}

\begin{figure}[t]
  \centering
  \includegraphics[width=\columnwidth]{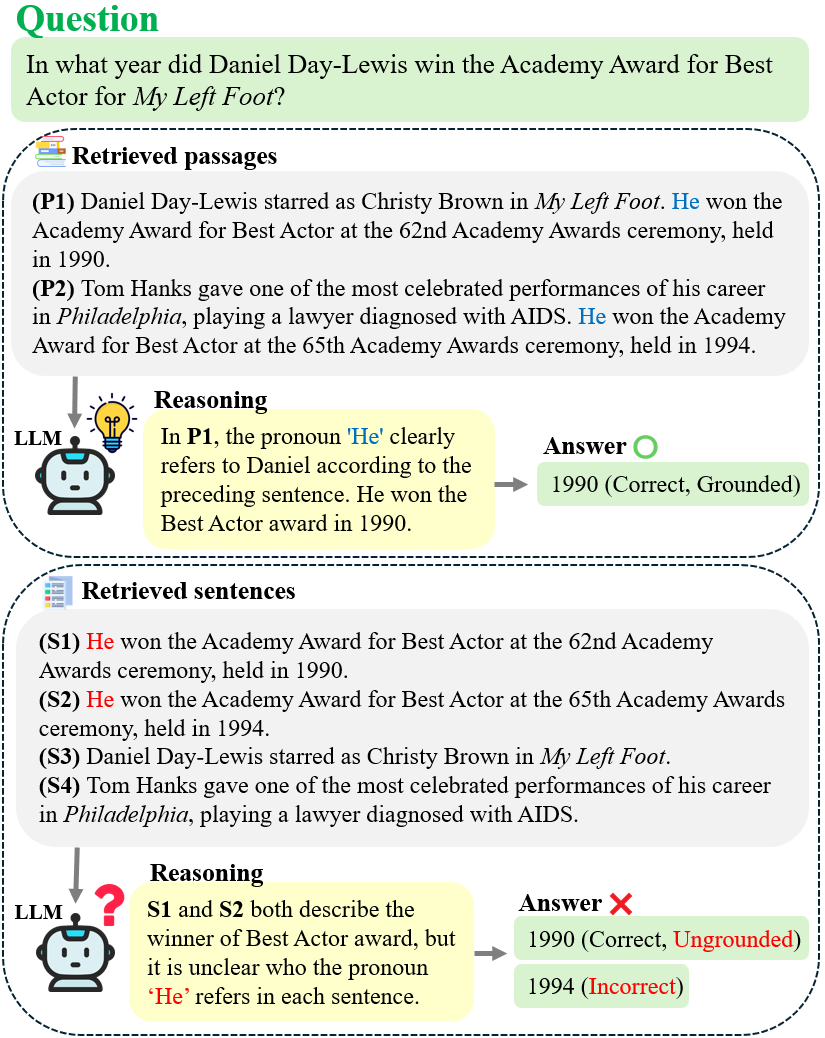}
\caption{The lost-in-generation problem. Even when correct evidence is
retrieved, sentence splitting leaves pronouns unresolvable, causing
incorrect generation.}
  \label{fig:lost_in_generation}
  \vspace{-8pt}
\end{figure}

Retrieval-Augmented Generation (RAG)~\cite{lewis2020retrieval} lets large
language models (LLMs) answer knowledge-intensive questions by drawing on
external corpora beyond their parametric knowledge. The setting is especially
demanding for multi-hop QA~\cite{yang2018hotpotqa,ho2020constructing,
trivedi2022musique}, where the evidence is scattered across documents, and RAG
systems commonly address this with two forms of decomposition: a complex query
is decomposed into a chain of sub-questions, and the corpus is decomposed into
finer-grained retrieval units such as sentences.

We observe that both forms of decomposition share an underlying vulnerability,
namely the loss of \textbf{entity} information. Each decomposition step can
remove a key entity that the surrounding context had made explicit, and this
loss causes failures at two different stages of the RAG pipeline.

\begin{figure}[t]
\centering
\small
\setlength{\tabcolsep}{5pt}
\renewcommand{\arraystretch}{1.4}
\begin{tabularx}{\linewidth}{@{}lX@{}}
\hline
\textbf{Sentence} &
  He won the Academy Award for Best Actor
  at the 62nd Academy Awards, held in 1990. \\
\hline
\textbf{Prefixed-sentence} &
  \textcolor{red}{Daniel Day-Lewis:}~He won
  the Academy Award for Best Actor at the 62nd Academy
  Awards, held in 1990. \\
\hline
\end{tabularx}
\vskip 4pt
\includegraphics[width=\linewidth]{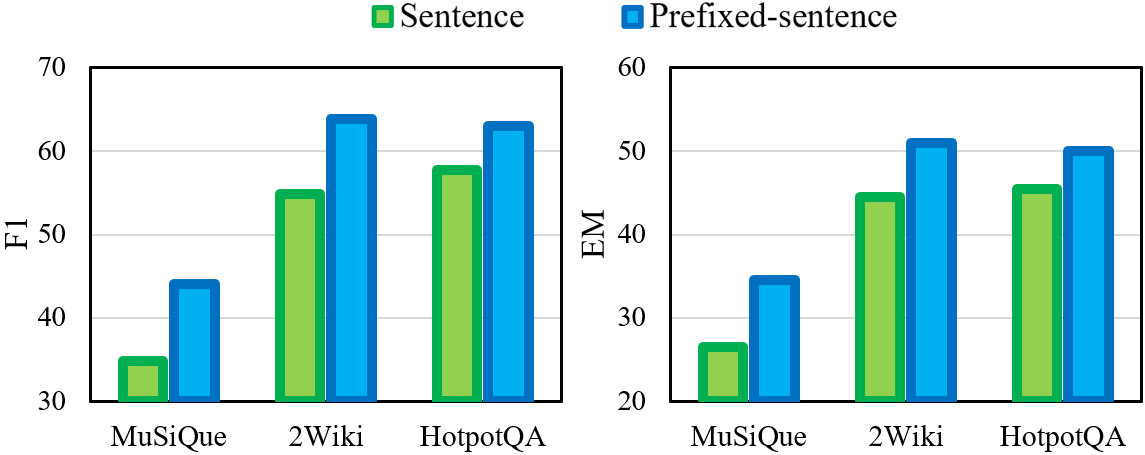}
\caption{(\textit{Top}) Example of Sentence and Prefixed-sentence.
\textcolor{red}{Daniel Day-Lewis} is the subject entity prefix that resolves pronoun ambiguity.
(\textit{Bottom}) Generation performance of Sentence vs.\ Prefixed-sentence
given the same retrieved sentence.}
\label{fig:prefixed_bar}
\vspace{-8pt}
\end{figure}

The first stage is retrieval. When a multi-hop question is broken into
sub-questions, a later sub-question often loses the entity resolved in the
preceding step and refers to it only through a vague pronoun or a
demonstrative, which degrades retrieval and disrupts the reasoning chain. This
problem was identified by ChainRAG~\cite{zhu2025mitigating} as
\textit{lost-in-retrieval} and addressed with a post-hoc step that rewrites a
sub-question to recover the missing entity. ChainRAG reports an average entity
recovery accuracy of 79.3\%, which leaves clear room for improvement and
motivates us to revisit the problem from a generation-time perspective.

The second stage is generation, and it is the failure mode we isolate.
Splitting a passage into sentences improves retrieval granularity, but it also
removes the context that disambiguates a sentence's pronouns. As illustrated in
Figure~\ref{fig:lost_in_generation}, once a passage is split into individual
sentences, a sentence such as ``\textit{He} won the Academy Award for Best
Actor at the 62nd Academy Awards ceremony, held in 1990'' becomes ambiguous,
and the LLM can no longer tell who ``He'' refers to. This phenomenon has been
studied as \textit{decontextualization} by \citet{choi2021decontextualization},
who address it by rewriting and re-indexing the corpus. Our contribution is to
isolate it as a distinct failure mode within the generation step of RAG, which
we call \textit{lost-in-generation}, where it persists after retrieval has
succeeded and is invisible to retrieval-side metrics. It exposes an assumption
RAG pipelines often make implicitly, that retrieving the correct evidence is
enough, since a correctly retrieved sentence is not always usable and retrieval
correctness does not ensure the evidence stays grounded once the LLM reads it.

This gap is not hypothetical. We run a controlled experiment in which the retrieved sentences are held fixed and the only variable is whether each sentence carries a subject entity prefix. Prepending the subject entity of the source passage improves exact match by +9.24, +9.08, and +5.18 on MuSiQue, 2Wiki, and HotpotQA (Figure~\ref{fig:prefixed_bar}), and since the evidence is identical in both conditions the gain isolates a generation-side failure rather than a retrieval one. We develop this into a full controlled study in Section~\ref{sec:exp_oracle}.

Motivated by these findings, we propose \textbf{Dual Entity Recovery RAG (DER-RAG)}, a RAG framework built on NaiveRAG with both decomposition strategies and organized around a single principle, that the entity grounding a piece of evidence should remain explicit from retrieval through to generation. DER-RAG addresses the two stages of entity
loss with one lightweight component each, requiring no graph construction, no
corpus modification, and no fine-tuning. For lost-in-retrieval, 
a two-way query decomposition generates each sub-question so
that the entity it depends on is already present, which removes the need for a
post-hoc rewriting step. For lost-in-generation, an LLM extracts the subject
entity of each passage offline, and this entity is prefixed to each retrieved 
sentence at generation stage. On MuSiQue, 2WikiMultiHopQA, and HotpotQA,
DER-RAG matches or exceeds strong baselines on average, including graph-based
methods that rely on offline graph construction. Our main contributions are as follows:
\begin{itemize}[leftmargin=1.2em, itemsep=2pt, topsep=2pt, partopsep=0pt, parsep=0pt]
 \item We connect two entity-loss problems that prior work studied separately, lost-in-retrieval in the retrieval step and the loss of pronoun grounding in the generation step, and isolate \textbf{lost-in-generation} as a distinct failure mode where corpus decomposition discards the entity that grounds a sentence.
  \item We propose \textbf{DER-RAG}, a lightweight framework that addresses both forms of entity loss without graph construction, corpus modification or fine-tuning.
  \item We evaluate DER-RAG on three multi-hop QA benchmarks, where it matches or exceeds strong baselines including graph-based RAGs, with a controlled experiment and ablations isolating each component's contribution.
\end{itemize}
\section{Related Work}
\label{sec:related}

\paragraph{Retrieval-augmented generation and multi-hop QA.}
RAG~\citep{lewis2020retrieval} grounds LLMs in external knowledge without
parameter updates, and a broad line of work improves it through iterative
retrieval~\citep{shao2023enhancing,trivedi2023interleaving},
reranking~\citep{glass2022re2g}, and context compression~\citep{wu2025lighter}.
Multi-hop QA~\citep{yang2018hotpotqa,ho2020constructing,trivedi2022musique} is
difficult because the evidence is scattered across documents, and RAG systems
commonly address this with two forms of decomposition. On the query side, a
complex question is decomposed into sub-questions~\citep{perez2020unsupervised,
khot2022decomposed, press2023measuring}, and methods such as
IRCoT~\citep{trivedi2023interleaving} interleave chain-of-thought reasoning
with retrieval at each step. On the corpus side, the retrieval unit ranges from
passages~\citep{karpukhin2020dense} to sentences~\citep{zhu2025mitigating} to
long grouped units~\citep{jiang2024longrag}. A prominent recent direction
pursues multi-hop performance through elaborate offline structures: graph-based
methods organise the corpus into hierarchical trees~\citep{sarthi2024raptor} or
entity graphs~\citep{edge2024local,gutierrez2024hipporag,li2024graphreader}
and retrieve by traversing them, which is effective but requires graph
construction whose cost grows with corpus size.
ChainRAG~\citep{zhu2025mitigating} takes a different view, identifying the
lost-in-retrieval problem in which entity underspecification in a sub-question
causes retrieval to fail regardless of how elaborate the downstream pipeline
is. DER-RAG follows this view, asking where entity information is lost along
the pipeline rather than where more structure can be added.

\paragraph{Decomposition depth and termination.}
A multi-step decomposition must decide how many sub-questions to generate. Much
prior work on these benchmarks assumes two-hop questions and decomposes
accordingly~\citep{zhu2025mitigating}, which is simple but does not generalise
to other reasoning depths. DeepRAG~\citep{guan2025deeprag} instead decides at
each step whether to continue, and because this decision is hard to make from
the query alone, it learns a termination policy through fine-tuning. DER-RAG
occupies neither end: it predicts the decomposition depth from the question
with a single prompted LLM call, which keeps the procedure general over
reasoning depth while requiring no fine-tuning and no training data.

\paragraph{Retrieval unit and decontextualization.}
The granularity of the retrieval unit is a central design choice in RAG, and
sentence-level retrieval improves precision over passage-level
retrieval~\citep{chen2024dense} while being widely
adopted~\citep{zhu2025mitigating}. A sentence taken out of its passage,
however, may no longer be interpretable on its own, since its pronouns and
definite descriptions lose their antecedents.
\citet{choi2021decontextualization} study this phenomenon as
\textit{decontextualization}, rewriting a sentence to be interpretable out of
context and re-indexing a corpus of such sentences. This line of work treats
interpretability as a static property of a sentence, fixed once at indexing
time. DER-RAG instead views referential failure as something a RAG pipeline
produces dynamically, in that decomposition breaks the grounding at query time,
after retrieval has already selected the evidence. We therefore address the
same phenomenon in a different place and by a different means, identifying it
as a distinct failure mode in the generation step of RAG, where it persists
after retrieval has succeeded and is invisible to retrieval-side metrics, and
we resolve it without rewriting or re-indexing the corpus. Rather than editing
the indexed text, DER-RAG prepends the subject entity of the source passage at
generation stage, which leaves the retrieval index unchanged and adds only a
single LLM call per passage offline.

\section{DER-RAG}
\label{sec:method}

\begin{figure*}[t]
  \centering
  \includegraphics[width=\textwidth]{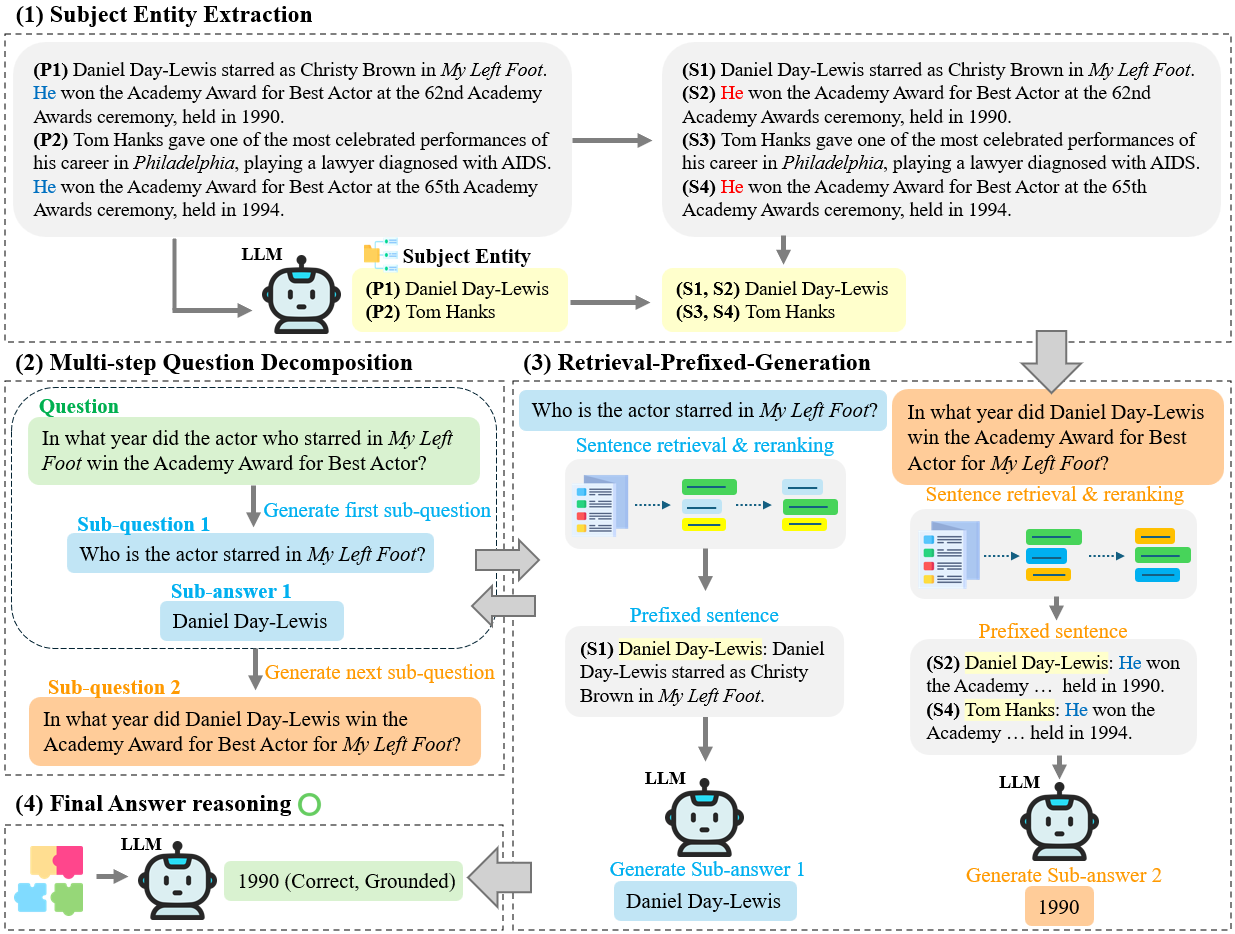}
  \caption{Overview of the DER-RAG pipeline.
  (1)~\textbf{Two-way query decomposition}: for non-comparison questions,
  sub-questions are generated sequentially, each conditioned on the answer
  to the preceding step.
  (2)~\textbf{Prefixed-sentence generation}: for each sub-question, supporting
  sentences are retrieved and prefixed with the subject entity of their source
  passage before being passed to the LLM.
  (3)~\textbf{Final answer reasoning}: the LLM aggregates prefixed sentences
  and sub-answers to produce the final answer.}
  \label{fig:DER-RAG_pipeline}
  \vspace{-5pt}
\end{figure*}

Multi-hop RAG systems rely on two complementary forms of decomposition, in that a complex query is split into a chain of sub-questions and the corpus is split into fine-grained retrieval units such as sentences. Both forms help the pipeline, since decomposition simplifies each retrieval step and finer units improve retrieval precision, but both also share one side effect, in that each decomposition step can strip away the entity information the surrounding context made explicit. DER-RAG is built on this standard pipeline, NaiveRAG with both decomposition strategies, and is organized around a single principle, that the entity grounding a piece of evidence should remain explicit from
retrieval through to generation.

Entity information is lost at two stages. The first is retrieval, where a later sub-question omits the entity resolved at the previous hop and refers to it only through a pronoun, leaving the retriever with no entity to match against (Section~\ref{sec:method_qd}). The second is generation, where a sentence
separated from its passage loses the context that licenses its pronouns, so the LLM cannot attribute the sentence to the correct entity even when it has been retrieved correctly (Section~\ref{sec:method_prefix}).

DER-RAG applies this principle with one lightweight component at each stage. At the query side, two-way query decomposition generates each sub-question so that the entity it depends on is already present, removing the need for a separate post-hoc rewriting step. At the generation side, prefixed-sentence generation restores the entity that sentence splitting discarded, attaching the subject entity of the source passage to each retrieved sentence before the LLM reads it. The two components are two instantiations of the same principle.

This design is lightweight by construction. Neither component requires graph construction, corpus modification or fine-tuning, and the only offline cost is a single LLM call per passage to extract its subject entity, which treats each passage independently and scales linearly with corpus size.
Figure~\ref{fig:DER-RAG_pipeline} illustrates the full pipeline.

\subsection{Two-Way Query Decomposition}
\label{sec:method_qd}

A sub-question generated during query decomposition can fail to retrieve
relevant passages whenever it lacks the intermediate entity resolved at the
previous hop. DER-RAG addresses this by routing each multi-hop question to one
of two decomposition paths, so that the intermediate entity is always explicit
in the sub-question that needs it.

\paragraph{Question type classification.}
Given a multi-hop question $q$, DER-RAG first classifies it as
\textit{comparison} or \textit{non-comparison} using a single LLM call that
returns a binary label. A question is labeled comparison when it explicitly
names two or more entities and asks them to be evaluated against a shared
criterion, as in ``Which came first, $X$ or $Y$?'' or ``Who is older, $A$ or
$B$?'', while all other questions, including bridge and compositional
questions, are labeled non-comparison. This distinction is highly reliable in
practice (Appendix~\ref{sec:classification_details}), so classification error
contributes negligibly to the pipeline, and the classification prompt is given
in Appendix~\ref{sec:prompts}.

For comparison questions, both targets are already named in the original
question, so DER-RAG generates all sub-questions in a single LLM call and each
sub-question inherits a named entity directly from $q$. No sub-question
therefore depends on an entity that has yet to be resolved, and this path also
avoids the sequential cost of multi-step decomposition. For non-comparison questions, including bridge and compositional questions, the entity needed to anchor a later sub-question is not known until the preceding sub-question has been answered, so DER-RAG generates sub-questions one at a time. At step $t$, the LLM receives the original question $q$ together with the history of previously generated sub-questions and their retrieved answers $\{(q_1, a_1), \ldots, (q_{t-1}, a_{t-1})\}$, and from these it generates the next sub-question $q_t$. Because $q_t$ is generated after $a_{t-1}$ has been retrieved, the intermediate entity resolved by $a_{t-1}$ is available, and DER-RAG instructs the LLM to substitute this entity directly into $q_t$ instead of referring to it with a pronoun. The decomposition prompts are given in Appendix~\ref{sec:prompts}.

\paragraph{Termination.}
A multi-step decomposition must decide how many sub-questions to generate. ChainRAG~\citep{zhu2025mitigating} and much prior work assume two-hop questions, while DeepRAG~\citep{guan2025deeprag} learns a termination policy through fine-tuning, at the cost of training data and additional compute. DER-RAG instead keeps this decision lightweight and train-free, decomposing
questions of arbitrary reasoning depth without a learned policy. After the
question type is classified, the LLM predicts, conditioned on the original
question, how many sub-questions the decomposition will require, and because
this prediction comes through a prompt instruction it needs no training data.
We report the predicted-count distribution in Appendix~\ref{sec:exp_steps}, and
the strong results in Section~\ref{sec:main_results} indicate that this simple
predictor is enough to support accurate multi-hop reasoning.

\subsection{Prefixed-Sentence Generation}
\label{sec:method_prefix}

Sentence-level retrieval improves precision, but a sentence read in isolation
often loses the entity that its pronouns and definite descriptions refer to, so
the LLM may fail to attribute a retrieved fact to the correct entity even when
the sentence itself is relevant and correctly retrieved. DER-RAG resolves this
by prefixing each retrieved sentence with the subject entity of its source
passage before the sentence is passed to the LLM.

\paragraph{Subject entity.}
For each passage $c$ in the corpus, we define its \textit{subject entity} $e_c$
as the central topic entity that the passage primarily describes. We obtain
$e_c$ with a single LLM call per passage, using the information-extraction
prompt in Figure~\ref{fig:prompt_entity}, which asks the LLM to return the one
entity the passage is primarily about as a concise noun phrase. This step
treats each passage independently, so it is trivially parallelizable. It is run
once, offline, and cached as passage metadata, so it adds no cost to online
inference, and the resulting $e_c$ is shared by every sentence split from $c$.

When a corpus provides document titles, the title can be used directly as
$e_c$. Titles are not always available, however, and even when present they
need not coincide with the passage's true subject, since a passage titled after
a single album in a discography is primarily about the artist rather than the
album. LLM extraction therefore serves as the general mechanism, with titles as
a readily available substitute. All DER-RAG experiments use extracted subject
entities, and we compare against title-based prefixes in
Appendix~\ref{sec:entity_extraction_quality}.

\paragraph{Prefixing.}
For each sub-question $q_t$, DER-RAG retrieves the top-$k$ sentences from the
indexed corpus using dense retrieval followed by cross-encoder reranking, and
retrieval operates on the original, unmodified sentences. At generation stage,
each retrieved sentence $s_i$ originating from passage $c$ is prefixed with the
subject entity $e_c$ of that passage to form
\begin{equation*}
  \tilde{s}_i = e_c \,\texttt{:}\; s_i .
\end{equation*}
The prefixed context $[\tilde{s}_1, \tilde{s}_2, \ldots, \tilde{s}_k]$ is then
passed to the LLM to generate the sub-answer $a_t$. Because each sentence now
carries an explicit statement of the entity it describes, the LLM can attribute
retrieved facts to the correct entity rather than to an unresolved pronoun.
This intervention operates only on the generation stage, so its effect is in
principle independent of retrieval quality, and we isolate this effect with a
retrieval-controlled experiment in Section~\ref{sec:exp_oracle} while also
ablating the prefix within the full pipeline in Section~\ref{sec:exp_ablation}.

Because the prefix is added at generation stage, the retrieval index is left
unchanged, so DER-RAG can be applied to an existing sentence-level RAG system
as a drop-in modification.

\begin{table*}[t]
\centering
\footnotesize
\setlength{\tabcolsep}{4pt}
\renewcommand{\arraystretch}{1.1}
\begin{tabular}{l p{3.5cm} cccccc}
\toprule
\multirow{2}{*}{\textbf{LLMs}} & \multirow{2}{*}{\textbf{Methods}}
  & \multicolumn{2}{c}{\textbf{MuSiQue}}
  & \multicolumn{2}{c}{\textbf{2Wiki}}
  & \multicolumn{2}{c}{\textbf{HotpotQA}} \\
\cmidrule(lr){3-4}\cmidrule(lr){5-6}\cmidrule(lr){7-8}
& & F1 & EM & F1 & EM & F1 & EM \\
\midrule
\multirow{10}{*}{\rotatebox[origin=c]{90}{\textbf{GPT-4o-mini}}}
  & Direct
    & $16.63_{\pm0.62}$ & $10.33_{\pm0.29}$
    & $32.07_{\pm1.15}$ & $26.50_{\pm1.00}$
    & $39.37_{\pm0.75}$ & $29.50_{\pm0.50}$ \\
  & NaiveRAG (passage)
    & $35.30_{\pm0.42}$ & $28.83_{\pm0.76}$
    & $51.99_{\pm0.56}$ & $41.83_{\pm0.76}$
    & $58.99_{\pm0.34}$ & $44.17_{\pm0.29}$ \\
  & NaiveRAG (sentence)
    & $44.39_{\pm0.54}$ & $35.33_{\pm0.76}$
    & $57.48_{\pm0.81}$ & $48.17_{\pm0.29}$
    & $61.36_{\pm0.71}$ & $47.83_{\pm0.76}$ \\
  & ITER-RETGEN
    & $42.62_{\pm0.51}$ & $32.17_{\pm0.76}$
    & $67.40_{\pm0.59}$ & $52.33_{\pm0.76}$
    & $59.68_{\pm1.30}$ & $48.33_{\pm1.26}$ \\
  & LongRAG
    & $47.02_{\pm1.49}$ & $36.83_{\pm0.76}$
    & $64.86_{\pm1.68}$ & $54.33_{\pm0.58}$
    & $\mathbf{66.59}_{\pm0.72}$ & $\mathbf{52.17}_{\pm0.76}$ \\
  & HippoRAG w/ IRCoT
    & $44.29_{\pm0.47}$ & $30.17_{\pm1.04}$
    & $63.07_{\pm0.99}$ & $50.50_{\pm1.50}$
    & $59.05_{\pm0.30}$ & $43.83_{\pm1.26}$ \\
  & ChainRAG (AnsInt)
    & $44.69_{\pm1.00}$ & $33.67_{\pm1.04}$
    & $63.16_{\pm2.42}$ & $53.67_{\pm2.02}$
    & $58.59_{\pm0.67}$ & $45.67_{\pm1.04}$ \\
  & ChainRAG (CxtInt)
    & $51.17_{\pm1.20}$ & $42.33_{\pm2.08}$
    & $62.15_{\pm2.13}$ & $52.67_{\pm2.57}$
    & $62.84_{\pm0.57}$ & $49.83_{\pm1.26}$ \\
\cmidrule(lr){2-8}
  & \cellcolor{cyan!10}DER-RAG (AnsInt)
    & \cellcolor{cyan!10}$\underline{56.46}_{\pm0.18}$ & \cellcolor{cyan!10}$\underline{46.50}_{\pm0.50}$
    & \cellcolor{cyan!10}$\mathbf{71.41}_{\pm0.22}$   & \cellcolor{cyan!10}$\mathbf{60.17}_{\pm0.29}$
    & \cellcolor{cyan!10}$57.49_{\pm0.96}$             & \cellcolor{cyan!10}$44.17_{\pm0.76}$ \\
  & \cellcolor{cyan!10}DER-RAG (CxtInt)
    & \cellcolor{cyan!10}$\mathbf{57.13}_{\pm0.42}$   & \cellcolor{cyan!10}$\mathbf{48.83}_{\pm0.58}$
    & \cellcolor{cyan!10}$\underline{70.67}_{\pm1.19}$ & \cellcolor{cyan!10}$\underline{59.33}_{\pm1.53}$
    & \cellcolor{cyan!10}$\underline{63.36}_{\pm0.06}$ & \cellcolor{cyan!10}$\underline{50.67}_{\pm0.29}$ \\
\midrule
\multirow{10}{*}{\rotatebox[origin=c]{90}{\textbf{Qwen2.5-72B}}}
  & Direct
    & $15.18_{\pm0.65}$ & $7.67_{\pm0.29}$
    & $31.34_{\pm0.61}$ & $26.17_{\pm0.58}$
    & $32.28_{\pm0.64}$ & $24.50_{\pm0.50}$ \\
  & NaiveRAG (passage)
    & $23.31_{\pm0.33}$ & $16.17_{\pm0.29}$
    & $48.91_{\pm0.43}$ & $39.00_{\pm0.00}$
    & $37.99_{\pm0.49}$ & $26.33_{\pm0.58}$ \\
  & NaiveRAG (sentence)
    & $41.64_{\pm0.22}$ & $30.00_{\pm0.00}$
    & $63.08_{\pm0.60}$ & $52.17_{\pm0.58}$
    & $56.98_{\pm0.10}$ & $43.67_{\pm0.29}$ \\
  & ITER-RETGEN
    & $40.53_{\pm0.67}$ & $30.83_{\pm0.76}$
    & $54.59_{\pm0.57}$ & $45.17_{\pm0.76}$
    & $59.26_{\pm0.85}$ & $46.33_{\pm0.58}$ \\
  & LongRAG
    & $40.69_{\pm0.40}$ & $29.50_{\pm0.50}$
    & $63.74_{\pm1.17}$ & $52.50_{\pm0.87}$
    & $60.10_{\pm0.25}$ & $46.67_{\pm0.29}$ \\
  & HippoRAG w/ IRCoT
    & $44.27_{\pm0.29}$ & $30.50_{\pm0.50}$
    & $64.46_{\pm0.53}$ & $53.00_{\pm1.32}$
    & $55.42_{\pm0.77}$ & $42.00_{\pm0.50}$ \\
  & ChainRAG (AnsInt)
    & $47.01_{\pm0.54}$ & $36.67_{\pm0.58}$
    & $64.69_{\pm0.98}$ & $54.50_{\pm1.32}$
    & $59.48_{\pm0.84}$ & $45.50_{\pm0.87}$ \\
  & ChainRAG (CxtInt)
    & $48.38_{\pm0.55}$ & $38.17_{\pm0.76}$
    & $64.62_{\pm0.99}$ & $54.17_{\pm1.26}$
    & $\mathbf{63.22}_{\pm0.75}$ & $\mathbf{50.67}_{\pm1.26}$ \\
\cmidrule(lr){2-8}
  & \cellcolor{cyan!10}DER-RAG (AnsInt)
    & \cellcolor{cyan!10}$\mathbf{55.00}_{\pm0.54}$   & \cellcolor{cyan!10}$\mathbf{43.17}_{\pm0.58}$
    & \cellcolor{cyan!10}$\underline{72.78}_{\pm0.35}$ & \cellcolor{cyan!10}$\mathbf{61.83}_{\pm0.29}$
    & \cellcolor{cyan!10}$59.65_{\pm1.35}$             & \cellcolor{cyan!10}$46.00_{\pm1.80}$ \\
  & \cellcolor{cyan!10}DER-RAG (CxtInt)
    & \cellcolor{cyan!10}$\underline{52.75}_{\pm0.90}$ & \cellcolor{cyan!10}$\underline{40.33}_{\pm0.58}$
    & \cellcolor{cyan!10}$\mathbf{73.15}_{\pm0.49}$   & \cellcolor{cyan!10}$\underline{60.67}_{\pm0.29}$
    & \cellcolor{cyan!10}$\underline{62.78}_{\pm0.76}$ & \cellcolor{cyan!10}$\underline{48.50}_{\pm0.87}$ \\
\bottomrule
\end{tabular}
\caption{Performance (\%) on MuSiQue, 2Wiki, and HotpotQA
         (mean $\pm$ std over 3 runs).
         \textbf{Bold}: best per LLM block; \underline{underline}: second best.}
\label{tab:main_results}
\vspace{-5pt}
\end{table*}

\subsection{Final Answer Reasoning}
\label{sec:method_final}
After all sub-questions have been answered, DER-RAG produces the final answer
using one of two integration strategies that follow~\citet{zhu2025mitigating},
where \textsc{AnsInt} derives the final answer from the accumulated
sub-questions and their answers and \textsc{CxtInt} derives it from the
retrieved sentence context. Under both strategies the LLM reasons over multiple
pieces of retrieved or accumulated evidence, so the same loss of entity
grounding can recur at the final-answer stage. DER-RAG therefore applies subject
entity prefixes at this stage as well, which keeps evidence grounded
consistently from retrieval through to the final answer. The
retrieval-controlled experiment in Section~\ref{sec:exp_oracle} covers both the
sub-answer and the final-answer stage.

\section{Experiments}
\label{sec:experiments}

\subsection{Experimental Setup}
\label{sec:setup}

We conduct experiments on MuSiQue~\citep{trivedi2022musique},
2WikiMultiHopQA~\citep{ho2020constructing}, and HotpotQA~\citep{yang2018hotpotqa},
following the 200-question data setting of~\citet{zhu2025mitigating}, and we
evaluate performance using F1 and exact match (EM). Dataset statistics are
given in Appendix~\ref{sec:dataset_stats}.

For a fair comparison we fix the embedding model (text-embedding-3-small) and the cross-encoder reranker as BGE-Reranker-large~\citep{chen2024bge}  across all systems. We compare DER-RAG against Direct (no retrieval), NaiveRAG (passage), NaiveRAG (sentence), ITER-RETGEN~\citep{shao2023enhancing}, LongRAG~\citep{jiang2024longrag},
HippoRAG w/~IRCoT~\citep{gutierrez2024hipporag,trivedi2023interleaving}, and ChainRAG~\citep{zhu2025mitigating}. Main experiments use GPT-4o-mini~\citep{achiam2023gpt} and Qwen2.5-72B~\citep{Yang2024Qwen25TR}, and ablation studies use GPT-4o-mini unless otherwise noted. We report the two answer integration strategies of ChainRAG~\citep{zhu2025mitigating}, where \textsc{AnsInt} derives the final answer from the accumulated sub-question and its own answer, and \textsc{CxtInt} derives it from the retrieved sentence context.

\subsection{Main Results}
\label{sec:main_results}

Table~\ref{tab:main_results} reports results on all three benchmarks for both
backbone LLMs. To ensure statistical reliability on the 200-question subsets,
every experiment is repeated three times and reported as mean $\pm$ standard
deviation. Averaged over the three benchmarks, DER-RAG matches or exceeds the strongest
baselines under both backbone LLMs, and on every individual dataset the better
of its two variants ranks first or second. The margins are clearest on MuSiQue
and 2WikiMultiHopQA, while on HotpotQA the methods fall within a narrow range,
which is consistent with its predominantly two-hop questions, where the room
for decomposition and entity recovery is smaller. Against the graph-based
HippoRAG w/ IRCoT, DER-RAG performs better on all three datasets without any
graph construction or offline preprocessing, which suggests that explicit query
decomposition with entity-aware prefixing is a lightweight alternative to
building offline structure. Among the baselines, sentence-level NaiveRAG
improves over its passage-level counterpart on all datasets, which confirms the
precision benefit of a finer retrieval unit, and LongRAG is strong on HotpotQA
for GPT-4o-mini but does not carry this advantage to MuSiQue and 2Wiki.

Advanced RAG methods generally surpass NaiveRAG in mean performance but also
show higher variance. ChainRAG in particular has large standard deviations, for
example ${\pm}2.42$ F1 on 2Wiki for GPT-4o-mini, which we attribute to noise
accumulating as the pipeline grows heavier, since more LLM calls and larger
retrieved contexts across steps raise the chance of error propagation. DER-RAG
instead reaches lower standard deviations than NaiveRAG in most settings, which
indicates that grounding the reasoning in a compact retrieved context yields
more consistent outputs, and this stability is clearest on MuSiQue and 2Wiki
where the accuracy gains are also largest.

We report DER-RAG under both answer integration strategies, \textsc{AnsInt} and
\textsc{CxtInt}, because entity loss can arise both when a sub-question is
answered and when the final answer is generated. DER-RAG matches or exceeds
strong baselines under both strategies on the large majority of datasets and
LLMs, which indicates that subject entity prefixing addresses the failure mode
across both points of answer synthesis.

\subsection{Lost-in-Generation under Controlled Retrieval}
\label{sec:exp_oracle}

This experiment tests the central claim that subject entity prefixing improves
generation independently of retrieval quality. To remove retrieval as a
confound, we bypass the retriever and build the context of each question from
all passages annotated for it, where the gold passages contain the answer and
the remaining passages serve as distractors. The correct evidence is therefore
guaranteed to be in the context in every condition. On this fixed context we
compare three ways of presenting the evidence to the LLM, the full passage, the
passage split into sentences, and the split sentences with a subject entity
prefix. The split sentences are reordered by the same reranker, applied
identically to all three evidence formats. Since the underlying passages are
identical in all three conditions, any difference in answer quality reflects
how the evidence is presented rather than what was retrieved. The fixed context
contains 8.6 to 11.1 passages per question on average, with details given in
Appendix~\ref{sec:dataset_stats}.

\begin{table}[t]
\centering
\small
\setlength{\tabcolsep}{3pt}
\renewcommand{\arraystretch}{1.2}
\begin{tabular}{lcccccc}
\toprule
\multirow{2}{*}{\textbf{Evidence Form}} &
  \multicolumn{2}{c}{\textbf{MuSiQue}} &
  \multicolumn{2}{c}{\textbf{2Wiki}} &
  \multicolumn{2}{c}{\textbf{HotpotQA}} \\
\cmidrule(lr){2-3}\cmidrule(lr){4-5}\cmidrule(lr){6-7}
& F1 & EM & F1 & EM & F1 & EM \\
\midrule
Passage
  & \textbf{49.18} & \textbf{39.5}
  & \textbf{65.12} & \textbf{53.5}
  & \underline{64.70} & \textbf{51.5} \\
Sentence
  & 46.24 & \underline{36.0}
  & 62.88 & 52.0
  & 62.02 & 48.5 \\
Prefixed sentence
  & \underline{47.85} & 35.5
  & \underline{64.55} & \textbf{53.5}
  & \textbf{65.24} & \underline{50.5} \\
\bottomrule
\end{tabular}
\caption{Lost-in-generation under controlled retrieval. The context is fixed to
the gold plus distractor passages, with no retriever run over the corpus.}
\label{tab:oracle}
\vspace{-7pt}
\end{table}

\begin{table}[t]
\centering
\small
\setlength{\tabcolsep}{3pt}
\renewcommand{\arraystretch}{1.15}
\begin{tabular}{lcccccc}
\toprule
\multirow{2}{*}{\textbf{QD Strategy}} &
  \multicolumn{2}{c}{\textbf{MuSiQue}} &
  \multicolumn{2}{c}{\textbf{2Wiki}} &
  \multicolumn{2}{c}{\textbf{HotpotQA}} \\
\cmidrule(lr){2-3}\cmidrule(lr){4-5}\cmidrule(lr){6-7}
& F1 & EM & F1 & EM & F1 & EM \\
\midrule
One-step QD
  & 39.41 & 31.50 & 59.19 & 49.00 & \textbf{65.65} & \textbf{52.50} \\
Post-hoc
  & \underline{45.43} & \underline{35.50} & \underline{60.00} & \underline{49.50} & \underline{65.28} & \underline{51.00} \\
Two-way QD
  & \textbf{49.87} & \textbf{41.00} & \textbf{64.43} & \textbf{53.00} & 63.63 & \underline{51.00} \\
\bottomrule
\end{tabular}
\caption{Comparison of query decomposition strategies on a NaiveRAG base.
Post-hoc follows ChainRAG~\citep{zhu2025mitigating}, and Two-way QD
is our method.}
\label{tab:qd_ablation}
\vspace{-7pt}
\end{table}

Table~\ref{tab:oracle} reports the result. Splitting a passage into sentences
lowers F1 on all three datasets, by 2.2 to 2.9 points. Because the retriever is
bypassed and the underlying passages are unchanged, this drop cannot come from
retrieval, which makes it a direct measurement of lost-in-generation, in that
splitting alone, by detaching a sentence from the context that resolves its
pronouns, degrades generation. Prefixing the subject entity recovers most
of this loss, improving over the unmodified sentence by 1.6 to 3.2 points and
bringing it back to roughly the full-passage level.

This also resolves an apparent tension with the ablation in
Section~\ref{sec:exp_ablation}, where the sentence outperforms the passage
under ordinary retrieval. The two orderings are consistent once retrieval and
generation are separated, since a finer unit helps retrieval~\citep{chen2024dense}
but hurts generation, and the plain sentence carries both effects, so its
retrieval gain is partly cancelled by lost-in-generation. The prefixed sentence
keeps the retrieval advantage of a fine unit, since the index is unchanged,
while repairing the generation side.

\subsection{Ablation Study}
\label{sec:exp_ablation}

DER-RAG modifies two parts of the standard pipeline, the query decomposition
strategy and the retrieval unit, so we ablate each on a common NaiveRAG base.
All ablation runs use the same embedding model, reranker, and evaluation setup
as the main results, so the numbers are directly comparable with
Table~\ref{tab:main_results}.

We first ablate the query decomposition strategy. Table~\ref{tab:qd_ablation}
compares three strategies on the same NaiveRAG base, where one-step QD generates
all sub-questions in advance without access to intermediate answers,
ChainRAG~\citep{zhu2025mitigating} adds a post-hoc rewriting step that recovers
omitted entities after generation, and two-way QD supplies the intermediate
entity while each sub-question is generated. On MuSiQue and 2Wiki, two-way QD
improves F1 by 4.4 points over post-hoc rewriting and by 10.5 and 5.2 over
one-step QD, which shows that supplying the entity during generation beats
recovering it afterward. On HotpotQA the three strategies perform within a
narrow range, with two-way QD slightly behind, consistent with its
predominantly two-hop questions where decomposition strategy matters little.
The advantage of two-way QD is therefore clearest on datasets that require
deeper or less uniform reasoning chains.

We next ablate the retrieval unit. Table~\ref{tab:unit_ablation} compares the
full passage, the unmodified sentence, and the prefixed sentence used by
DER-RAG on the same NaiveRAG base. The unmodified sentence improves over the
passage, confirming the precision benefit of a finer unit, and prefixing then
adds a further gain over the unmodified sentence, by 3.0, 9.5, and 4.0 F1 points
on MuSiQue, 2Wiki, and HotpotQA. Since the prefixed and unmodified sentences are
retrieved identically and differ only by the prefix, this gain isolates entity
grounding at generation time rather than any retrieval effect, and the
comparison against the passage condition shows it comes from restoring the
missing entity, not from supplying more context.
\begin{table}[t]
\centering
\small
\setlength{\tabcolsep}{3pt}
\renewcommand{\arraystretch}{1.2}
\begin{tabular}{lcccccc}
\toprule
\multirow{2}{*}{\textbf{Retrieval Unit}} &
  \multicolumn{2}{c}{\textbf{MuSiQue}} &
  \multicolumn{2}{c}{\textbf{2Wiki}} &
  \multicolumn{2}{c}{\textbf{HotpotQA}} \\
\cmidrule(lr){2-3}\cmidrule(lr){4-5}\cmidrule(lr){6-7}
& F1 & EM & F1 & EM & F1 & EM \\
\midrule
Passage
  & 34.89 & 28.0
  & 51.51 & 41.0
  & 58.74 & 44.0 \\
Sentence
  & \underline{44.68} & \textbf{36.0}
  & \underline{56.97} & \underline{48.0}
  & \underline{60.54} & \underline{47.0} \\
Prefixed sentence
  & \textbf{47.72} & \underline{35.5}
  & \textbf{66.47} & \textbf{55.5}
  & \textbf{64.51} & \textbf{52.5} \\
\bottomrule
\end{tabular}
\caption{Retrieval unit ablation on a NaiveRAG base. Prefixed sentence is the
unit used by DER-RAG.}
\label{tab:unit_ablation}
\vspace{-7pt}
\end{table}

\begin{table}[t]
\centering
\small
\setlength{\tabcolsep}{3pt}
\renewcommand{\arraystretch}{1.2}
\begin{tabular}{l l @{\hspace{10pt}} cc cc cc}
\toprule
\multirow{2}{*}{\textbf{Methods}} & \multirow{2}{*}{}
  & \multicolumn{2}{c}{\textbf{MuSiQue}}
  & \multicolumn{2}{c}{\textbf{2Wiki}}
  & \multicolumn{2}{c}{\textbf{HotpotQA}} \\
\cmidrule(lr){3-4} \cmidrule(lr){5-6} \cmidrule(lr){7-8}
 & & F1 & EM & F1 & EM & F1 & EM \\
\midrule
\multirow{2}{*}{\makecell[l]{DER-RAG}}
    & AnsInt  & 56.25 & 46             & \underline{71.66} & \underline{60}   & 58.02          & 45 \\
  & CxtInt  & \underline{56.76}   & \textbf{48.5}  & \textbf{72.04} & \textbf{61} & \underline{63.41} & \underline{51} \\
\cmidrule(lr){1-8}
\multirow{2}{*}{\makecell[l]{DER-RAG\\(w/o routing)}}
    & AnsInt  & 56.14 & \underline{46.5} & 67.01 & 55.5          & 59.47          & 46 \\
  & CxtInt  & \textbf{57.41} & \textbf{48.5}    & 68.06 & 56.5 & \textbf{65.86} & \textbf{52} \\
\bottomrule
\end{tabular}
\caption{Effect of Two-Way Routing ($k$=10).}
\label{tab:routing}
\vspace{-7pt}
\end{table}
\begin{table}[t]
\centering
\small
\setlength{\tabcolsep}{3pt}
\renewcommand{\arraystretch}{1.2}
\begin{tabular}{l l @{\hspace{10pt}} cc cc cc}
\toprule
\multirow{2}{*}{\textbf{Methods}} & \multirow{2}{*}{}
  & \multicolumn{2}{c}{\textbf{MuSiQue}}
  & \multicolumn{2}{c}{\textbf{2Wiki}}
  & \multicolumn{2}{c}{\textbf{HotpotQA}} \\
\cmidrule(lr){3-4} \cmidrule(lr){5-6} \cmidrule(lr){7-8}
 & & F1 & EM & F1 & EM & F1 & EM \\
\midrule
\multirow{2}{*}{\makecell[l]{DER-RAG}}
    & AnsInt  & \underline{56.25} & \underline{46}   & \underline{71.66} & \underline{60} & 58.02          & 45 \\
  & CxtInt  & \textbf{56.76}   & \textbf{48.5}  & \textbf{72.04} & \textbf{61} & \textbf{63.41} & \textbf{51} \\
\cmidrule(lr){1-8}
\multirow{2}{*}{\makecell[l]{DER-RAG\\(w/o prefix)}}
    & AnsInt  & 54.94 & 44.5             & 63.15          & 52.5          & 53.07          & 42 \\
  & CxtInt  & 49.45 & 38.5    & 65.70 & 54.5 & \underline{62.18} & \underline{49.5} \\
\bottomrule
\end{tabular}
\caption{Effect of Sentence Prefixing ($k$=10).}
\label{tab:prefix_ablation}
\vspace{-7pt}
\end{table}

\subsection{Effect of Two-Way Routing}
\label{sec:exp_routing}

DER-RAG routes comparison questions to one-step decomposition and all other
questions to multi-step decomposition. This experiment isolates the
contribution of that routing by comparing the full two-way design against a
variant that removes the routing and applies multi-step decomposition to every
question, including comparison questions. All other components, the retriever,
the prefixing, and the generation setup, are held fixed, so any difference
reflects only the routing.

Table~\ref{tab:routing} compares the full two-way routing design against 
a variant that applies multi-step decomposition to every question without routing.
The results vary across datasets depending on the proportion of comparison-type questions
(see Table~\ref{tab:qtype} for the distribution of question types per dataset).

On MuSiQue, which contains no comparison questions, removing the routing yields 
virtually no change in performance, confirming that the routing mechanism has 
no adverse effect when comparison questions are absent.
On 2Wiki, which has the highest proportion of comparison questions (82 out of 200), 
the no-routing variant shows a notable performance drop, 
demonstrating the effectiveness of the two-way routing in redirecting 
comparison questions to the more appropriate one-step decomposition path.
On HotpotQA, the performance gap between the two variants is relatively small. 
We attribute this to the comparatively lower accuracy on comparison-type 
questions in HotpotQA, which diminishes the marginal benefit of routing.
Taken together, these findings validate the utility of the two-way routing 
strategy: it provides meaningful gains precisely in settings where 
comparison questions are frequent and answerable.

\subsection{Effect of Sentence Prefixing}
\label{sec:exp_prefix}
DER-RAG prefixes each retrieved sentence with the subject entity of its
source passage before the sentence is passed to the LLM. To isolate the prefix effect, 
we compare DER-RAG against a variant that strips the prefix and passes each retrieved sentence
to the LLM unmodified. Table~\ref{tab:prefix_ablation} shows that prefixing improves performance
on all three datasets and under both integration strategies, with no
exception. This is a broader pattern than the one observed for two-way
routing in Section~\ref{sec:exp_routing}, whose benefit was concentrated on 2Wiki. Because
lost-in-generation arises whenever a passage is split into sentences,
regardless of question type or reasoning depth, the prefix that repairs it
helps consistently across all settings rather than in a narrow subset of
cases.

\subsection{Top-$k$ Study}
\label{sec:topk}

\begin{figure}[t]
  \centering
  \includegraphics[width=\linewidth]{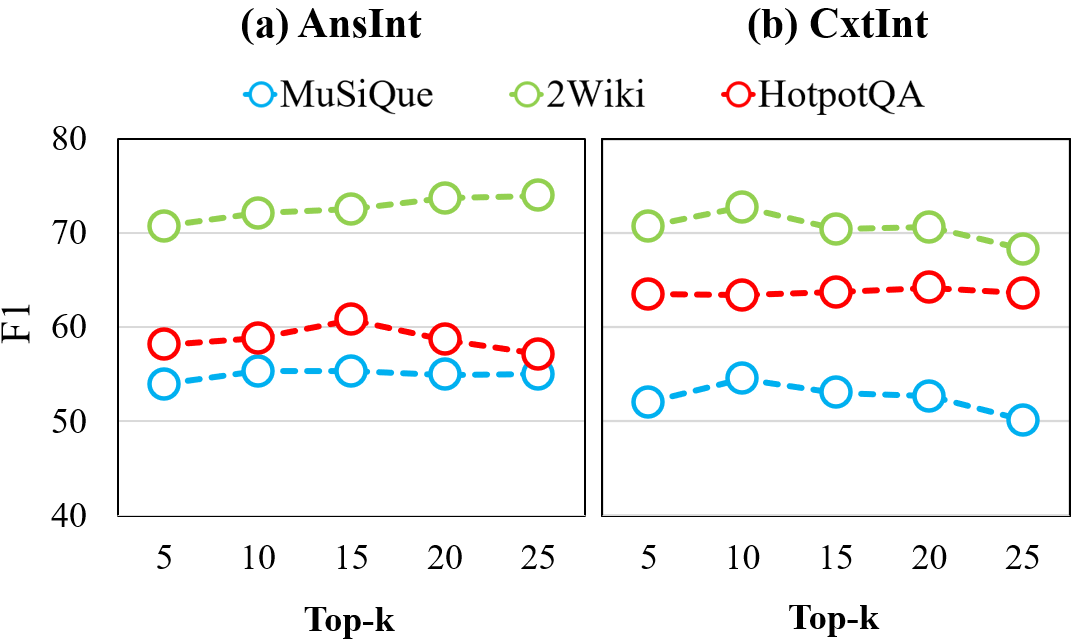}
  \caption{F1 as a function of Top-$k$, for AnsInt (left) and CxtInt (right)
  across all three datasets.}
  \label{fig:topk_line}
\end{figure}

Figure~\ref{fig:topk_line} plots F1 as a function of the number of retrieved
sentences $k$, swept from 5 to 25.
Across both integration types and all three datasets, performance remains 
relatively stable.
A notable exception is 2Wiki under \textsc{AnsInt}, where F1 shows a gradual increase 
as $k$ grows, suggesting that broader retrieval coverage benefits 
this dataset's multi-hop structure.
However, beyond $k{=}10$, larger values tend to yield diminishing returns 
or a slight decline, an effect that is more pronounced under textsc{CxtInt}.
We attribute this to the fact that \textsc{CxtInt} concatenates all retrieved contexts 
from each sub-question and passes them jointly to the generation stage, 
causing the input to grow substantially with $k$ and introducing greater context noise.
We therefore adopt $k{=}10$ as the default.

\subsection{Online Efficiency}
\label{sec:efficiency}
We compare DER-RAG and ChainRAG on online inference cost along two dimensions,
the number of LLM calls (Table~\ref{tab:llm_calls}) and the number of retrieved
context sentences (Table~\ref{tab:retrieved_sents}). DER-RAG issues a comparable
number of LLM calls to ChainRAG, slightly more on each dataset, since two-way
query decomposition follows a short chain of sub-questions. It operates,
however, on far fewer retrieved sentences, around 25 to 33 per question against
67 to 104 for ChainRAG, because subject entity prefixing restores entity
grounding directly rather than retrieving additional sentences to supply the
missing context. DER-RAG therefore reaches its accuracy with a substantially
smaller generation context at a similar number of LLM calls.

\begin{table}[t]
\centering
\small
\renewcommand{\arraystretch}{1.15}
\begin{tabular}{lccc}
\toprule
\textbf{} &
\textbf{MuSiQue} & \textbf{2Wiki} & \textbf{HotpotQA} \\
\midrule
ChainRAG & 8.98 & 7.52 & 8.00 \\
DER-RAG   & 9.63 & 8.00 & 9.11 \\
\bottomrule
\end{tabular}%
\caption{Comparison of LLM calls (GPT-4o-mini).}
\label{tab:llm_calls}
\vspace{-7pt}
\end{table}
\begin{table}[t]
\centering
\small
\renewcommand{\arraystretch}{1.15}
\begin{tabular}{lccc}
\toprule
\textbf{} &
\textbf{MuSiQue} & \textbf{2Wiki} & \textbf{HotpotQA} \\
\midrule
ChainRAG & 103.84 & 67.24 & 75.36 \\
DER-RAG   & 33.10 & 24.94 & 30.34 \\
\bottomrule
\end{tabular}%
\caption{Comparison of retrieved sentences N.}
\label{tab:retrieved_sents}
\vspace{-7pt}
\end{table}

\section{Conclusion}
\label{sec:conclusion}
We revisited entity loss in multi-hop RAG and showed that it breaks the
pipeline at two stages. Lost-in-retrieval, identified in prior work, occurs
when a sub-question drops the entity resolved at the previous hop.
Lost-in-generation, which we isolate as a distinct failure mode in the
generation step, occurs when sentence splitting detaches a sentence from the
context that grounds its pronouns, and a retrieval-controlled experiment shows
that it degrades answers even when the gold evidence is fixed. We proposed
DER-RAG, which keeps the grounding entity explicit across both stages with two
lightweight components and requires no graph construction, no corpus modification, 
and no fine-tuning. Across three multi-hop QA benchmarks, DER-RAG
matches or exceeds strong baselines, including graph-based methods, which
suggests that addressing entity loss directly is a worthwhile alternative to
building heavier retrieval structures.

\section{Limitations} 
Three limitations of DER-RAG follow from its core design. First, DER-RAG assigns a single subject entity to each passage and reuses it for every sentence and every query. Our analysis in Appendix~\ref{sec:entity_extraction_quality} shows
that this is often too rigid, since the most useful subject entity is not a
fixed expression but varies in granularity, and which form helps most can
depend on the query being answered. A passage-level entity also cannot resolve
every coreference chain within a passage, because some sentences refer to
entities other than the main subject. Extracting finer, query-aware entities
is a natural next step. Second, DER-RAG decides the decomposition depth once, 
right after the question type classification. This keeps the method train-free 
and is enough to support accurate reasoning on the benchmarks we study, but a 
policy that revisits the decision at every step could adapt more finely to 
questions whose reasoning depth is hard to predict in advance. Making such a 
per-step policy train-free, rather than relying on fine-tuning as current 
approaches do, is a natural direction for future work.
Third, our evaluation is conducted entirely on Wikipedia-derived corpora 
(MuSiQue, 2WikiMultiHopQA, and HotpotQA). Empirical validation on real-world, 
domain-specific corpora is needed before DER-RAG's results can be assumed to 
generalize beyond these three benchmarks.

\section*{Acknowledgements}
The Authors were supported by the Institute of Information \& Communications Technology Planning \& Evaluation(IITP) grants funded by the Korea government(MSIT) (RS-2025-25442824, AI Star Fellowship Program(Ulsan National Institute of Science and Technology)), (No. RS-2020-II201336, Artificial Intelligence graduate school support(UNIST)), and (No. RS-2026-25527734, Hyper-scale Industrial AI Research Support (R\&D) Program, Development of industry-specified multimodal hyperscale foundation model based technologies for manufacturing, predictive maintenance, and process optimization) and by Korea Institute for Advancement of Technology (KIAT) grant funded by the Korea Government (MOTIE) (P0029947, Company-friendly industry-academia convergence promotion support project(Ulsan)).
\bibliography{custom}

\clearpage
\appendix

\section{Dataset Statistics and Implementation Details}
\label{sec:dataset_stats}
Table~\ref{tab:dataset_stats} presents statistics for the three benchmarks used
in our evaluation. Each dataset comprises 200 questions, and the corpus
associated with each question consists of multiple passages of varying length.
MuSiQue and HotpotQA tend to have longer passages, with 43.69 and 47.93
sentences per passage on average, while 2WikiMultiHopQA passages are shorter at
22.75 sentences on average. This difference in passage density is relevant to
lost-in-generation, since denser passages produce more pronoun-bearing
sentences once they are split, which increases the need for subject entity
prefixing.
\begin{table}[h]
\centering
\small
\begin{tabular}{lccc}
\toprule
 & \textbf{MuSiQue} & \textbf{2Wiki} & \textbf{HotpotQA} \\
\midrule
No. of Samples      & 200    & 200   & 200   \\
No. of Passages     & 2,219  & 1,986 & 1,722 \\
Avg. Sents          & 43.69  & 22.75 & 47.93 \\
\bottomrule
\end{tabular}
\caption{Dataset statistics. Avg.\ Sents: average number of sentences per passage.}
\label{tab:dataset_stats}
\end{table}
For a fair comparison across all methods we fix the embedding model
(text-embedding-3-small) and the cross-encoder reranker (BGE-Reranker-large)
throughout. Sentence splitting uses the "en\_core\_web\_sm" model 
from spaCy library, and all methods operate on the same preprocessed corpus. For
ITER-RETGEN we follow the original settings and report results at the third
iteration, where performance plateaus. LongRAG and HippoRAG are evaluated with
their publicly released code under default configurations, with only the
backbone LLM and embedding model replaced for a unified comparison, where
LongRAG applies word-count-based chunking with a chunk size of 200 and HippoRAG
operates on unsegmented passages. For DER-RAG, subject entities are extracted
offline in a single preprocessing pass and cached, which adds no latency to
online inference; Table~\ref{tab:extraction-cost} reports the resulting cost.
All ablation experiments use GPT-4o-mini unless otherwise noted.
 
\begin{table}[h]
\centering
\small
\begin{tabular}{lcccc}
\toprule
\textbf{Corpus} & \textbf{LLM calls} & \textbf{Cost} & \textbf{Time} \\
\midrule
MuSiQue  & 2,219 & \$0.49 & 41.2m \\
2Wiki    & 1,986 & \$0.24 & 36.4m \\
HotpotQA & 1,722 & \$0.41 & 33.8m \\
\midrule
Total    & 5,927 & \$1.15 & 111.4m \\
\bottomrule
\end{tabular}
\caption{One-time offline subject-entity extraction cost
(GPT-4o-mini).}
\label{tab:extraction-cost}
\end{table}

The cost per LLM call varies noticeably across corpora, and this variation
does not track the number of calls alone. As shown in
Table~\ref{tab:dataset_stats}, the average number of sentences per passage
differs substantially across corpora, and since each call encodes one full
passage, a passage with more sentences consumes more input tokens per call.
2Wiki has the fewest sentences per passage among the three corpora, so its
per-call cost is correspondingly lower, which is why 2Wiki incurs the
lowest total extraction cost despite having a comparable number of passages
to MuSiQue and HotpotQA.

For the controlled experiment in Section~\ref{sec:exp_oracle}, the context of each question is fixed to its gold passages together with distractor passages drawn from the same pool, so that the correct evidence is always present. The combined passages are
then split into sentences and ordered by the same reranker, and the
prefixed-sentence condition additionally prepends the subject entity of each
source passage. Table~\ref{tab:oracle_context_stats} reports the resulting
number of passages per question, which is used in Section~\ref{sec:exp_oracle}. 
It averages 8.6 to 11.1 across the three datasets and varies from as few as 2 to as many as 20, 
so the fixed context still contains a substantial amount of non-supporting material.

All datasets (MuSiQue, 2WikiMultiHopQA, HotpotQA) and models used in this work are publicly available and were used in accordance with their respective licenses and intended research use.
\begin{table}[h]
\centering
\small
\begin{tabular}{lccc}
\toprule
 & \textbf{MuSiQue} & \textbf{2Wiki} & \textbf{HotpotQA} \\
\midrule
Mean & 11.09 & 9.93 & 8.61 \\
Min  & 3     & 6    & 2    \\
Max  & 20    & 10   & 10   \\
\bottomrule
\end{tabular}
\caption{Number of passages (gold plus distractor) in the fixed context used
for the controlled experiment in Section~\ref{sec:exp_oracle}.}
\label{tab:oracle_context_stats}
\end{table}

\section{Question Type Classification Details}
\label{sec:classification_details}

Table~\ref{tab:qtype} reports the accuracy of the LLM-based binary classifier
that routes each question to its decomposition path. The horizontal A/B labels 
denote the ground-truth question types (actual labels), while the vertical A/B 
labels denote the classifier’s predicted question types (predicted labels).
The classifier reaches over 99\% accuracy on all three datasets, with 100\% on 
MuSiQue, 99.5\% on 2WikiMultiHopQA, and 99\% on HotpotQA. The few misclassified 
cases are predominantly bridge questions that name two entities in their surface 
form, which leads the classifier to label them as comparison questions. Given this
near-perfect accuracy, classification error contributes negligibly to overall
pipeline performance. The classification prompt is shown in
Figure~\ref{fig:prompt_classify}. As shown below, this is a simple prompt 
guided by a small set of few-shot examples.

\definecolor{tblue}{RGB}{181,212,244}

\begin{table}[t]
\centering
\small
\setlength{\extrarowheight}{3pt}
\begin{tabular}{l cc cc cc}
\toprule
& \multicolumn{2}{c}{\textbf{MuSiQue}}
& \multicolumn{2}{c}{\textbf{2Wiki}}
& \multicolumn{2}{c}{\textbf{HotpotQA}} \\
\cmidrule(lr){2-3} \cmidrule(lr){4-5} \cmidrule(lr){6-7}
& \textbf{A} & \textbf{B}
& \textbf{A} & \textbf{B}
& \textbf{A} & \textbf{B} \\
\midrule
A & \cellcolor{tblue}0   & 0
  & \cellcolor{tblue}82  & 1
  & \cellcolor{tblue}42  & 2   \\
B & 0   & \cellcolor{tblue}200
  & 0   & \cellcolor{tblue}117
  & 0   & \cellcolor{tblue}156 \\
\midrule
EM & \multicolumn{2}{c}{100}
   & \multicolumn{2}{c}{99.5}
   & \multicolumn{2}{c}{99} \\
\bottomrule
\end{tabular}
\caption{Question type classification results using GPT-4o-mini.
A: comparison type, B: non-comparison type.
Diagonal cells indicate correct classifications.}
\label{tab:qtype}
\end{table}
\begin{figure}[h]
\begin{tcolorbox}[
  enhanced,
  colback=white,
  colframe=black!70,
  arc=5pt,
  boxrule=1.2pt,
  left=8pt, right=8pt, top=8pt, bottom=8pt,
  fontupper=\small,
  title={\textbf{Classify Comparative Question}},
  colbacktitle=black!75,
  coltitle=white,
  fonttitle=\small,
  toptitle=0.1pt,
  bottomtitle=0.1pt,
]
You are a classifier that determines whether a question is Comparative or Not.

\vspace{10pt}
A question is Comparative if and only if:
\begin{enumerate}[label=\arabic*., leftmargin=*, itemsep=1pt]
  \item It involves TWO or more explicitly named entities (not derived via reasoning).
  \item The final answer requires a direct comparison between those entities.
  \item The answer is one of: yes/no, one of the named entities, or a shared/differing attribute.
\end{enumerate}

A question is Not Comparative if:
\begin{enumerate}[label=\arabic*., leftmargin=*, itemsep=1pt]
  \item One entity must be derived first before the comparison can be made. (bridge)
  \item Multiple entities/items appear in the question but serve as clues, not comparison targets.
\end{enumerate}

Example: \{few-shots\}
\end{tcolorbox}
\caption{Prompt for classifying whether a question is comparative or not.}
\label{fig:prompt_classify}
\end{figure}

\section{Decomposition Depth}
\label{sec:exp_steps}
\begin{table}[h]
\centering
\small
\setlength{\tabcolsep}{10pt}
\renewcommand{\arraystretch}{1.25}
\begin{tabular}{c ccc}
\toprule
 & \textbf{MuSiQue} & \textbf{2Wiki} & \textbf{HotpotQA} \\
\midrule
0 &   0 &  82 &  42 \\
1 &  13 &   1 &  12 \\
2 & 187 & 117 & 146 \\
\midrule
Total & 200 & 200 & 200 \\
\bottomrule
\end{tabular}
\caption{Distribution of predicted sub-question counts per dataset.
         Depth~0 indicates comparison-type questions routed to
         one-step decomposition, which are excluded from multi-step QD.}
\label{tab:decomp_depth}
\end{table}
This section reports analyses of decomposition depth on the multi-step decomposition.
DER-RAG predicts the number of sub-questions after question type classification, rather 
than fixing it in advance (Section~\ref{sec:method_qd}). We record the predicted count 
for every question and report its distribution per dataset, which shows that, although 
the procedure is not restricted to a fixed depth, it terminates within a small
number of steps for the large majority of questions on these benchmarks.
Table~\ref{tab:decomp_depth} shows the distribution of predicted sub-question
counts across the three datasets.
Questions assigned zero sub-questions are comparison-type questions that the
two-way routing directs to one-step decomposition; they are excluded from
multi-step QD entirely.
Among the remaining questions, the vast majority are assigned exactly two
sub-questions across all datasets, with only a small fraction receiving one,
confirming that the benchmarks are dominated by two-hop reasoning chains.
No question is assigned three or more sub-questions, indicating that the
predicted depth aligns well with the actual reasoning requirements of these
benchmarks and that the procedure terminates compactly without over-decomposing.

\section{Portability of Two-Way Query Decomposition}
\label{sec:qd_portability}

The query decomposition ablation in Section~\ref{sec:exp_ablation} measures
two-way QD on a NaiveRAG base. Here we additionally test whether the same
decomposition transfers to a different pipeline, by replacing the post-hoc
rewriting step of ChainRAG~\citep{zhu2025mitigating} with two-way QD while
leaving the rest of ChainRAG unchanged. Table~\ref{tab:chainrag_results}
reports the result.

\definecolor{rowgray}{RGB}{242,242,242}
\begin{table*}[t]
\centering
\renewcommand{\arraystretch}{1.2}
\setlength{\tabcolsep}{6pt}
\small
\begin{tabular}{llc cc cc cc}
\toprule
\multirow{2}{*}{\textbf{Method}} &
\multirow{2}{*}{\textbf{QD}} &
\multirow{2}{*}{} &
\multicolumn{2}{c}{\textbf{MuSiQue}} &
\multicolumn{2}{c}{\textbf{2Wiki}} &
\multicolumn{2}{c}{\textbf{HotpotQA}} \\
\cmidrule(lr){4-5}\cmidrule(lr){6-7}\cmidrule(lr){8-9}
& & & F1 & EM & F1 & EM & F1 & EM \\
\midrule
\multirow{4}{*}{ChainRAG}
  & \multirow{2}{*}{Post-hoc rewriting}
    & AnsInt & 50.54 & 37.0 & 62.55 & 52.0 & 60.73 & 46.0 \\
  & & CxtInt & 47.87 & 38.5 & 56.54 & 50.5 & \underline{64.59} & \underline{50.0} \\
\cmidrule(lr){2-9}
  & \multirow{2}{*}{Two-way}
    & AnsInt & \underline{54.00} & \underline{43.5} & \textbf{68.57} & \textbf{58.0} & 61.20 & 46.5 \\
  & & CxtInt & \textbf{54.05} & \textbf{45.5} & \underline{63.41} & \underline{53.5} & \textbf{66.14} & \textbf{52.0} \\
\bottomrule
\end{tabular}
\caption{Performance comparison of replacing ChainRAG's post-hoc rewriting with our two-way QD.}
\label{tab:chainrag_results}
\end{table*}
As shown in Table~\ref{tab:chainrag_results}, replacing post-hoc rewriting 
with two-way QD consistently improves or matches performance across all three datasets, 
demonstrating that two-way QD resolves lost-in-retrieval more effectively than post-hoc rewriting.
Also, this result indicates that the proposed decomposition strategy is not tied to a specific pipeline 
and can serve as a portable drop-in replacement within existing RAG frameworks.

\section{Title vs.\ Subject Entity}
\label{sec:entity_extraction_quality}

Table~\ref{tab:subject_title_similarity} reports the similarity between
LLM-extracted subject entities and original passage titles, measured by
token-level F1 and exact match. F1 scores of 82 to 86 across all three datasets
indicate that the extracted entities closely match the passage titles in most
cases, and the gap between F1 and EM reflects partial matches where the LLM
produces a slightly different surface form of the correct entity, such as
``Daniel Day-Lewis'' against ``Daniel Day Lewis''.

\begin{table}[h]
\centering
\small
\begin{tabular}{lccc}
\toprule
\textbf{} & \textbf{MuSiQue} & \textbf{2Wiki} & \textbf{HotpotQA} \\
\midrule
F1 & 86.06 & 82.31 & 82.76 \\
EM & 66.97 & 57.30 & 57.38 \\
\bottomrule
\end{tabular}
\caption{Similarity of subject entities extracted by GPT-4o-mini to the
original passage titles.}\label{tab:subject_title_similarity}
\end{table}
Table~\ref{tab:title_vs_subject} directly compares the two prefix strategies in
the full DER-RAG pipeline at $k{=}10$. Subject entity prefixing is comparable to
or better than title prefixing across most settings, and the gain is largest on
2WikiMultiHopQA under \textsc{CxtInt}, where subject entity prefixing improves F1 from
68.84 to 72.04. This is consistent with our case study, in which LLM-extracted
entities more precisely identify the central topic of a passage, particularly
when the title refers to a sub-entity rather than the primary subject. Title
prefixing is marginally higher on MuSiQue under \textsc{AnsInt}, which suggests that the
effect of prefix precision varies with both the dataset and the answer
integration strategy. We observe the same pattern on a NaiveRAG base, where
title and subject entity prefixes yield comparable performance across the three
datasets, which indicates that the choice between the two prefixes has a
limited effect on accuracy and that extracted subject entities can be used as
the general mechanism without a performance cost.
\begin{table}[h]
\centering
\renewcommand{\arraystretch}{1.15}
\setlength{\tabcolsep}{4pt}
\footnotesize
\begin{tabular}{lc cc cc cc}
\toprule
\multirow{2}{*}{\textbf{Prefix}} &
\multirow{2}{*}{} &
\multicolumn{2}{c}{\textbf{MuSiQue}} &
\multicolumn{2}{c}{\textbf{2Wiki}} &
\multicolumn{2}{c}{\textbf{HotpotQA}} \\
\cmidrule(lr){3-4}\cmidrule(lr){5-6}\cmidrule(lr){7-8}
& & F1 & EM & F1 & EM & F1 & EM \\
\midrule
\multirow{2}{*}{Title}
  & AnsInt & \textbf{56.82} & 45.5          & 71.56          & \textbf{61.0} & 56.98          & 43.5          \\
  & CxtInt & 55.48          & \underline{46.0} & 68.84       & 57.5          & \underline{62.98} & \underline{49.0} \\
\midrule
\multirow{2}{*}{Subject}
  & AnsInt & 56.25          & \underline{46.0} & \underline{71.66} & \underline{60.0} & 58.02     & 45.0          \\
  & CxtInt & \underline{56.76} & \textbf{48.5} & \textbf{72.04} & \textbf{61.0} & \textbf{63.41} & \textbf{51.0} \\
\bottomrule
\end{tabular}
\caption{Comparison of prefix strategies (title vs.\ subject entity) in DER-RAG ($k{=}10$).}
\label{tab:title_vs_subject}
\end{table}
To understand these results we conducted a human evaluation that categorises
each passage into three cases, defined in Table~\ref{tab:comparison_study},
where each case was identified by human annotators.
Table~\ref{tab:case_distribution} reports the resulting distribution. Case~1,
in which only the title is correct, accounts for fewer than 3\% of passages
on every dataset, which confirms that this failure mode is relatively rare.

\definecolor{tblue}{RGB}{181,212,244}
\definecolor{sgreen}{RGB}{192,221,151}
\definecolor{casebg}{RGB}{245,245,243}
\definecolor{okgreen}{RGB}{34,139,34}
\definecolor{errred}{RGB}{220,50,50}

\newcommand{\hlT}[1]{\colorbox{tblue}{#1}}
\newcommand{\hlS}[1]{\colorbox{sgreen}{#1}}

\begin{table*}[h]
\centering
\small
\setlength{\extrarowheight}{3pt}
\begin{tabular}{>{\raggedright\arraybackslash}p{8.0cm}
                >{\raggedright\arraybackslash}p{2.2cm}
                >{\raggedright\arraybackslash}p{4.2cm}}
\toprule
\textbf{Passage} & \textbf{Check} & \textbf{Analysis} \\
\midrule

\multicolumn{3}{l}{\cellcolor{casebg}\textbf{Case 1}} \\
Countess \hlT{Albina du Boisrouvray} is a former journalist and film
producer who has become a global philanthropist\ldots
She is the founder of \hlS{FXB International},
a non-governmental organization established in memory of her son.
& \makecell[tl]{{\color{okgreen}$\checkmark$}~\hlT{title} \\[4pt]
               {\color{errred}$\times$}~\hlS{subject}}
& Title correctly points to the person the passage is about, but subject captures the organization she founded — a related but distinct entity. \\[4pt]

\midrule
\multicolumn{3}{l}{\cellcolor{casebg}\textbf{Case 2}} \\
\hlS{Thordis Markusdottir}, known as \hlT{Stokkseyrar-Disa}
(1668--1728), was an Icelandic magician (Galdrmaster).
She is known for her alleged magical powers and is the subject
of at least ten different folk sagas.
& \makecell[tl]{{\color{okgreen}$\checkmark$}~\hlT{title} \\[4pt]
               {\color{okgreen}$\checkmark$}~\hlS{subject}}
& The passage opens by giving her real name and immediately states she is "known as" the title — both names refer to the same person. \\[4pt]

\midrule
\multicolumn{3}{l}{\cellcolor{casebg}\textbf{Case 3}} \\
\hlS{Aziza Mustafa Zadeh} (born 1969) is an Azerbaijani singer,
pianist, and composer who plays a fusion of jazz and mugham\ldots
Discography: Aziza Mustafa Zadeh (1991), Always (1993),
Dance of Fire (1995), \hlT{Jazziza} (1997)\ldots
& \makecell[tl]{{\color{errred}$\times$}~\hlT{title} \\[4pt]
               {\color{okgreen}$\checkmark$}~\hlS{subject}}
& The title is just one album listed in the discography section, subject correctly identifies the artist the passage describes. \\

\bottomrule
\end{tabular}
\caption{Case study of subject entity extraction (EM $<$ 1).}
\label{tab:comparison_study}
\end{table*}
\begin{table}[t]
\centering
\small
\setlength{\tabcolsep}{4.5pt}
\renewcommand{\arraystretch}{1.2}
\resizebox{\columnwidth}{!}{%
\begin{tabular}{lccc}
\toprule
\textbf{} &
  \textbf{MuSiQue} &
  \textbf{2Wiki} &
  \textbf{HotpotQA} \\
\midrule
Case 1 & 53~~(2.4\%)   & 46~~(2.3\%)   & 43~~(2.5\%)   \\
Case 2 & 2,155~(97.1\%) & 1,915~(96.4\%) & 1,665~(96.7\%) \\
Case 3 & 11~~(0.5\%)   & 25~~(1.3\%)   & 14~~(0.8\%)   \\
\midrule
Total  & 2,219 & 1,986 & 1,722 \\
\bottomrule
\end{tabular}%
}
\caption{Distribution of subject entity extraction cases per dataset.
Case~1: title correct, subject incorrect.
Case~2: both title and subject correct.
Case~3: title incorrect, subject correct (subject outperforms title).}
\label{tab:case_distribution}
\end{table}

To measure the downstream effect of these errors rather than relying on
intrinsic annotation alone, we replace the subject entity with the correct
title for every Case~1 passage and rerun DER-RAG, holding retrieval,
decomposition, and generation otherwise unchanged.
Table~\ref{tab:case1_corrected} reports the result.

Correcting all identified extraction errors changes F1 by at most 1.04
points and EM by at most 2.0 points, with no consistent directional effect.
This is expected given the rarity of Case~1: since it accounts for only
2.3--2.5\% of passages across the three datasets (Table~\ref{tab:case_distribution}),
even a complete correction of every such error can shift only a small
fraction of the corpus, which bounds how much aggregate performance can move
regardless of how the individual errors are resolved. This indicates that
the extraction-error rate has limited aggregate impact on final-answer
performance in the evaluated setting.

Two further patterns emerged from the evaluation. First, as Case~3 shows, the
extracted entity sometimes corrects an inaccurate title, since when a title
refers to a sub-entity such as a single album in a discography passage, the LLM
identifies the artist as the primary subject and yields a more informative
prefix than the title. Second, even in Case~2, where both the title and the
subject entity are judged correct, extraction often provides a complementary
advantage, because many titles are abbreviated forms such as a shortened place
name or a partial person name, whereas the LLM tends to produce a fuller
surface form that is better suited as a prefix. This suggests that extracted
entities can be a better prefix even when the title is technically accurate.

These observations also point to a deeper question, namely that the most
appropriate subject entity for a passage is not a single fixed expression but
can take several valid forms that differ in specificity, and which form is most
effective as a prefix may further depend on the query being answered. One
direction for future work is therefore to identify a small set of candidate
entities at varying granularity per passage and select the most query-relevant
one at retrieval time.

\begin{table}[t]
\centering
\setlength{\tabcolsep}{3pt}
\renewcommand{\arraystretch}{1.2}
\resizebox{\linewidth}{!}{%
\begin{tabular}{l l @{\hspace{10pt}} cc cc cc}
\toprule
\multirow{2}{*}{\textbf{Subject Entity}} & \multirow{2}{*}{}
  & \multicolumn{2}{c}{\textbf{MuSiQue}}
  & \multicolumn{2}{c}{\textbf{2Wiki}}
  & \multicolumn{2}{c}{\textbf{HotpotQA}} \\
\cmidrule(lr){3-4} \cmidrule(lr){5-6} \cmidrule(lr){7-8}
 & & F1 & EM & F1 & EM & F1 & EM \\
\midrule
\multirow{2}{*}{Original}
    & AnsInt  & 56.25 & \underline{46.0} & \underline{71.66} & \underline{60.0} & 58.02 & 45.0 \\
  & CxtInt  & \underline{56.76} & \textbf{48.5} & \textbf{72.04} & \textbf{61.0} & \textbf{63.41} & \textbf{51.0} \\
\cmidrule(lr){1-8}
\multirow{2}{*}{Modified}
    & AnsInt  & 55.29 & 45.5 & 70.62 & \underline{60.0} & 57.97 & 44.5 \\
  & CxtInt  & \textbf{56.79} & \textbf{48.5} & 71.32 & 59.0 & \underline{62.46} & \underline{49.5} \\
\bottomrule
\end{tabular}%
}
\caption{DER-RAG performance (GPT-4o-mini) on different subject entities.
Original: LLM-extracted subject entities. Modified: Based on the Original, Case~1 subject entities
replaced by the correct title.}
\label{tab:case1_corrected}
\vspace{-7pt}
\end{table}

\section{Case Study}
\label{sec:case_study}

Table~\ref{tab:case_study} traces a multi-hop QA example through three RAG
methods to show how the two entity-loss problems arise and how DER-RAG resolves
them.

For NaiveRAG w/~QD, the first sub-question correctly identifies the bridging
nursery rhyme, but because one-step QD generates all sub-questions at once
without access to intermediate answers, the second sub-question is formed
without substituting the resolved entity, which leaves the retriever unable to
identify the correct context and produces an incorrect answer. For ChainRAG,
post-hoc rewriting recovers the missing entity and issues a well-formed second
sub-question, which fixes retrieval, but the retrieved sentences all read ``It
has a Roud Folk Song Index number of \ldots'' with no indication of which
nursery rhyme each sentence describes, so the LLM cannot resolve the implicit
subject and again answers incorrectly. For DER-RAG, the second sub-question is
the same well-formed one as in ChainRAG, and each retrieved sentence is then
prefixed with the subject entity of its passage, so the LLM identifies the
sentence belonging to the target nursery rhyme and produces the correct answer.
This example shows that lost-in-retrieval and lost-in-generation are
independent failure modes, since even when retrieval succeeds the absence of
entity grounding in the generated context can still lead to a wrong answer, and
fixing only one problem is insufficient while the other remains.

\section{Prompts}
\label{sec:prompts}
DER-RAG implements each of its decision points, question type
classification, decomposition depth prediction, and sub-question
generation, through prompt engineering rather than fine-tuning or
additional model components. This is precisely what keeps the method
lightweight, but it also means that a non-trivial share of DER-RAG's
performance rests on the quality of the underlying prompts rather than on
learned parameters. We therefore validated each prompt-based component
empirically rather than treating it as a black box. The prompts themselves are
reported below for completeness and reproducibility.

DER-RAG uses an LLM with task-specific prompts at three points: subject entity extraction, question type classification, and sub-question generation. All prompts include few-shot demonstrations, omitted here for brevity, and are applied without any task-specific fine-tuning. The subject entity extraction prompt is shown in Figure~\ref{fig:prompt_entity} and the classification prompt in Figure~\ref{fig:prompt_classify}.

Figures~\ref{fig:prompt_comparative} to~\ref{fig:prompt_last_subq} present the sub-question generation prompts. Figure~\ref{fig:prompt_comparative} is the one-step decomposition prompt for comparison questions, which splits the question into two parallel sub-questions, one per named entity. Figure~\ref{fig:prompt_first_subq} is the prompt for the first sub-question in the multi-step chain for non-comparison questions.
Figure~\ref{fig:prompt_last_subq} is the prompt for the final sub-question, which substitutes the entity resolved at the preceding step directly into the
sub-question wording.

\section{Use of AI Assistants}
\label{sec:ai_use}
AI assistants were used for writing assistance, including grammar
checking, sentence-level editing, and improving the clarity and conciseness of
the text. All technical content, experimental results, and
claims in this paper are the authors' own.

\begin{figure}[H]
\begin{tcolorbox}[
  enhanced,
  colback=white,
  colframe=black!70,
  arc=5pt,
  boxrule=1.2pt,
  left=8pt, right=8pt, top=8pt, bottom=8pt,
  fontupper=\small,
  title={\textbf{Extract subject-entity}},
  colbacktitle=black!75,
  coltitle=white,
  fonttitle=\small,
  toptitle=0.1pt,
  bottomtitle=0.1pt,
]
You are an information extraction assistant.

\vspace{10pt}
Your task is to identify the single primary subject entity of a given paragraph.
The subject entity is the main object, person, place, concept, or event that the paragraph is primarily describing.\\[1pt]

\textbf{Rules:}
\begin{itemize}[label=--, leftmargin=*, nosep, itemsep=3pt]
  \item Output only the entity name, nothing else.
  \item Use a concise noun phrase (e.g., \textit{examples}).
  \item Do not output sentences, explanations, or punctuation.
  \item If multiple entities are present, choose the most central one.
\end{itemize}
\end{tcolorbox}
\caption{Prompt for extracting subject entity of a paragraph.}
\label{fig:prompt_entity}
\end{figure}

\begin{table*}[t]
\centering
\footnotesize
\setlength{\tabcolsep}{4pt}
\renewcommand{\arraystretch}{1.3}

\begin{tabular}{
  >{\raggedright\arraybackslash}p{4.5em}
  >{\raggedright\arraybackslash}p{13.0em}
  >{\raggedright\arraybackslash}p{26.0em}
  >{\centering\arraybackslash}p{2.5em}
}
\toprule
\multicolumn{4}{p{\dimexpr\linewidth-8pt}}{%
  \textbf{Question:} What was the Roud Folk Song Index of the nursery rhyme
  inspiring \textit{What Are Little Girls Made Of?}%
}\\[2pt]
\midrule
\textbf{Method} & \textbf{Sub-question 1} & \textbf{Sub-question 2} & \textbf{Answer} \\
\midrule

NaiveRAG w/ QD
&
\textbf{Q:} What is the nursery rhyme that inspired
\textit{`What Are Little Girls Made Of'?}

\smallskip
\textbf{Context:}
...the title is taken from the nursery rhyme, \textcolor{blue}{``What Are Little Boys Made Of?''}

\smallskip
\textbf{Answer:}
\textcolor{blue}{``What Are Little Boys Made Of?''}
&
\textbf{Q:} What is the Roud Folk Song Index of
\textcolor{red}{this nursery rhyme?}

\smallskip
\textbf{Context:}

\textbullet~ The nursery rhyme has a Roud Folk Song Index number of \textcolor{red}{19800}.

\textbullet~ It has a Roud Folk Song Index number of 7734.

\textbullet~ It has a Roud Folk Song Index number of 12986.

\textbullet~ It has a Roud Folk Song Index number of 821.

\smallskip
\textbf{Answer:} \textcolor{red}{19800}
&
\textcolor{red}{19800}
\\
\midrule

ChainRAG
&
\textbf{Q:} What is the nursery rhyme inspiring
\textit{What Are Little Girls Made Of?}

\smallskip
\textbf{Context:}
...the title is taken from the nursery rhyme, \textcolor{blue}{``What Are Little Boys Made Of?''}

\smallskip
\textbf{Answer:}
\textcolor{blue}{What Are Little Boys Made Of?}
&
\textbf{Q:} What was the Roud Folk Song Index of
\textcolor{blue}{What Are Little Boys Made Of?}

\smallskip
\textbf{Context:}

\textbullet~ It has a Roud Folk Song Index number of 19800.

\textbullet~ It has a Roud Folk Song Index number of 7734.

\textbullet~ It has a Roud Folk Song Index number of \textcolor{red}{12986}.

\textbullet~ It has a Roud Folk Song Index number of 821.

\smallskip
\textbf{Answer:} \textcolor{red}{12986}
&
\textcolor{red}{12986}
\\
\midrule

DER-RAG (ours)
&
\textbf{Q:} What is the nursery rhyme inspiring
\textit{What Are Little Girls Made Of?}

\smallskip
\textbf{Context:}
...the title is taken from the nursery rhyme, \textcolor{blue}{``What Are Little Boys Made Of?''}

\smallskip
\textbf{Answer:}
\textcolor{blue}{What Are Little Boys Made Of?}
&
\textbf{Q:} What was the Roud Folk Song Index of
\textcolor{blue}{What Are Little Boys Made Of?}

\smallskip
\textbf{Context:}

\textbullet~ \textcolor{green!80!black}{Tweedledum and Tweedledee}: ... Song Index number of 19800.

\textbullet~ \textcolor{green!80!black}{Three Little Kittens}: ... Song Index number of 7734.

\textbullet~ \textcolor{green!80!black}{A Wise Old Owl}: ... Song Index number of 12986.

\textbullet~ \textcolor{green!80!black}{What Are Little Boys Made Of?}:
... Song Index number of \textcolor{blue}{821}.

\smallskip
\textbf{Answer:} \textcolor{blue}{821}
&
\textcolor{blue}{821}
\\
\bottomrule
\end{tabular}
\caption{%
  RAG process comparison on a HotpotQA dataset.
  \textcolor{blue}{Blue} = correct/relevant; \textcolor{red}{red} = incorrect;
  \textcolor{green!80!black}{green} = subject entities prefixed to retrieved sentences.}
\label{tab:case_study}
\end{table*}
\begin{figure*}[t]
\begin{tcolorbox}[
  enhanced,
  colback=white,
  colframe=black!70,
  arc=5pt,
  boxrule=1.2pt,
  left=8pt, right=8pt, top=8pt, bottom=8pt,
  fontupper=\small,
  title={\textbf{Single-step QD}},
  colbacktitle=black!75,
  coltitle=white,
  fonttitle=\small,
  toptitle=0.1pt,
  bottomtitle=0.1pt,
]
You are a helpful AI assistant that decomposes a comparative question.

\vspace{8pt}
\textbf{Guidelines:}
\begin{enumerate}[label=\arabic*., leftmargin=*, itemsep=0.5pt]
  \item Output exactly 2 sub-questions as a JSON array (and nothing else).
  \item Each sub-question should extract the relevant attribute or fact for one of the two compared entities.
  \item Stay close to the original question wording by reusing it.
  \item If the original question specifies what type of entity is being compared (e.g., film, book, person, song), always include that type word in each sub-question.
\end{enumerate}

\vspace{4pt}
Example: \{few-shots\}
\end{tcolorbox}
\caption{Prompt for decomposing a comparative question into two sub-questions.}
\label{fig:prompt_comparative}
\end{figure*}
\begin{figure*}[t]
\begin{tcolorbox}[
  enhanced,
  colback=white,
  colframe=black!70,
  arc=5pt,
  boxrule=1.2pt,
  left=8pt, right=8pt, top=8pt, bottom=8pt,
  fontupper=\small,
  title={\textbf{Multi-step QD (First Sub-question)}},
  colbacktitle=black!75,
  coltitle=white,
  fonttitle=\small,
  toptitle=0.1pt,
  bottomtitle=0.1pt,
]
You are a helpful AI assistant that generates only one sub-question for multi-hop reasoning.

\vspace{8pt}
\textbf{Goal:} Produce the single best first sub-question.

\vspace{8pt}
\textbf{What is a sub-question:}
\begin{enumerate}[label=--, leftmargin=*, itemsep=0.5pt]
  \item Ask for the missing link (an intermediate entity/value).
  \item Do not ask for the final answer to the original question.
  \item Stay close to the original question wording by reusing key phrases/entities.
\end{enumerate}

\textbf{Hard rules:}
\begin{enumerate}[label=\arabic*., leftmargin=*, itemsep=0.5pt]
  \item Output exactly one sub-question.
  \item Avoid vague or generic prompts (e.g., ``Who is A?'') unless the original question truly requires identifying A.
  \item Do not use pronouns or demonstratives that rely on previous answers (forbidden: it, this, that, they, these, those, he, she, his, her, its, their).
  \item Make the subject explicit using entities already in the original question.
\end{enumerate}

\vspace{4pt}
Example: \{few-shots\}
\end{tcolorbox}
\caption{Prompt for generating the first sub-question for non-comparative question.}
\label{fig:prompt_first_subq}
\end{figure*}
\begin{figure*}[t]
\begin{tcolorbox}[
  enhanced,
  colback=white,
  colframe=black!70,
  arc=5pt,
  boxrule=1.2pt,
  left=8pt, right=8pt, top=8pt, bottom=8pt,
  fontupper=\small,
  title={\textbf{Multi-step QD (Last Sub-question)}},
  colbacktitle=black!75,
  coltitle=white,
  fonttitle=\small,
  toptitle=0.1pt,
  bottomtitle=0.1pt,
]
You are a helpful AI assistant that generates the last sub-question for the original question.

\vspace{8pt}
\textbf{Input:}
\begin{enumerate}[label=--, leftmargin=*, itemsep=0.5pt]
  \item Original question: user's final goal
  \item Current sub-question: question just answered
  \item Current answer: answer to the current sub-question
\end{enumerate}

\textbf{Task:}
\begin{enumerate}[label=--, leftmargin=*, itemsep=0.5pt]
  \item Using the current answer as a known fact, generate exactly one final sub-question that directly leads to the answer of the original question.
  \item This sub-question must be the last step; its answer should directly resolve the original question.
\end{enumerate}

\textbf{Hard rules:}
\begin{enumerate}[label=\arabic*., leftmargin=*, itemsep=0.5pt]
  \item Output only one question in JSON (no lists, no explanations).
  \item Do not use pronouns or demonstratives that require context (forbidden: it, this, that, they, these, those, he, she, his, her, its, their).
  \item Substitute the relevant entity from the current answer directly into the sub-question.
  \item Stay close to the original question wording by reusing it.
  \item If the original question contains family relations (e.g., ``maternal grandfather'', ``father-in-law''), determine which relation step has been resolved by the current sub-question, then ask only the remaining next step.
\end{enumerate}

\vspace{4pt}
Example: \{few-shots\}
\end{tcolorbox}
\caption{Prompt for generating the last sub-question for non-comparative question.}
\label{fig:prompt_last_subq}
\end{figure*}

\end{document}